%% file: _main.tex
\input{_constants}
\arxiv 

\pdfoutput=1
\documentclass[10pt,twocolumn,letterpaper]{article}
\input{cvpr_header}

\begin{document}
\title{Diffusion-Augmented Coreset Expansion for Scalable Dataset Distillation}
\author{\authorBlock}
\maketitle

\begin{abstract}
    With the rapid scaling of neural networks, data storage and communication demands have intensified. Dataset distillation has emerged as a promising solution, condensing information from extensive datasets into a compact set of synthetic samples by solving a bilevel optimization problem. However, current methods face challenges in computational efficiency, particularly with high-resolution data and complex architectures. Recently, knowledge-distillation-based dataset condensation approaches have made this process more computationally feasible. Yet, with the recent developments of generative foundation models, there is now an opportunity to achieve even greater compression, enhance the quality of distilled data, and introduce valuable diversity into the data representation. In this work, we propose a two-stage solution. First, we compress the dataset by selecting only the most informative patches to form a coreset. Next, we leverage a generative foundation model to dynamically expand this compressed set in real-time—enhancing the resolution of these patches and introducing controlled variability to the coreset. Our extensive experiments demonstrate the robustness and efficiency of our approach across a range of dataset distillation benchmarks. We demonstrate a significant improvement of over 10\% compared to the state-of-the-art on several large-scale dataset distillation benchmarks. The code will be released soon. 
\end{abstract}

\vspace{-0.5cm}

\section{Introduction}
With the rapid advancement of deep learning, the scale of neural networks and the datasets required to train them have expanded dramatically, introducing significant computational challenges. One promising approach to mitigate these demands is to explore the potential of ``small data,'' a research direction introduced by \citet{wang2018dataset} and known as dataset distillation. Dataset distillation focuses on synthesizing a compact yet highly informative dataset from the original large-scale data, allowing models trained on this smaller set to achieve performance comparable to those trained on the full dataset \cite{yu2023dataset}. By reducing training overhead, storage, and communication requirements while preserving the essential knowledge of the larger dataset, dataset distillation offers transformative potential across multiple areas of machine learning research. Its impact is particularly significant in applications that 1) require repeated training over large datasets, as seen in neural architecture search \cite{such2020generative}, 2) depend on constrained memory storage, such as memory replay in continual learning \cite{sangermano2022sample}, and 3) involve knowledge sharing and communication across distributed machine learning agents, such as in federated learning \cite{xiong2023feddm}.


Dataset distillation is classically formulated as a bilevel optimization problem, where the inner loop trains a model on a synthesized dataset, and the outer loop adjusts this synthetic dataset to maximize the model's performance on the original large-scale dataset. However, this approach presents two significant challenges: 1) it is computationally and memory intensive, as the outer optimization requires backpropagation through the entire unrolled computation graph of the model's training process in the inner loop; and 2) it often leads to synthetic images with spurious, non-realistic features due to overfitting to the specific architecture used during optimization, which limits generalization across architectures \cite{cazenavette2023generalizing}. To address the former challenge and scale up computation, researchers have proposed methods such as gradient matching \cite{zhao2021dataset}, which aligns the gradients of synthetic and original data to improve scalability, and training trajectory matching \cite{cazenavette2022dataset}, which matches training trajectories between models trained on synthetic and original datasets to enhance distillation efficiency. To address the latter problem, recent studies emphasize the importance of realism in distilled data for achieving cross-architecture generalizability \cite{cazenavette2023generalizing,yin2024squeeze,sun2024diversity,shao2024generalized}, showing that incorporating generative priors and enhancing the diversity and realism of synthetic datasets can significantly improve the generalization capabilities of models trained on distilled data.

Recently, \citet{sun2024diversity} introduced the Realistic, Diverse, and Efficient Dataset Distillation (RDED) method, an optimization-free approach that achieves high-resolution and large-scale image dataset distillation by emphasizing the realism and diversity of the distilled images. RDED selects a diverse set of informative patches directly cropped from the original data and combines these patches into new images to form the synthetic dataset. To guide this selection, RDED uses a ``teacher" model trained on the large-scale dataset to identify informative patches and assigns a soft label for each patch. A ``student" model is then trained on these informative patches along with their corresponding soft labels, effectively employing knowledge distillation for dataset distillation.

In RDED, we observe a trade-off between patch diversity and realism for a fixed compression budget, i.e., images per class (IPC). Increasing diversity requires reducing the size of patches to fit more patches within the limited pixel space. However, decreasing patch sizes also reduces their realism, as downsampling acts as a low-pass filter, causing a loss of fine-grained details. In this context, we pose two questions: 1) Is it possible to pack more patches into a finite pixel space without sacrificing realism? and 2) Can we enhance diversity within a fixed number of patches? Following the RDED framework \cite{sun2024diversity}, to increase the number of patches without sacrificing realism, super-resolution techniques can be used to enhance low-resolution patches back to high-resolution quality. Additionally, diversity within a fixed number of patches can be achieved through realistic augmentations that preserve the natural image manifold. We show that modern Latent Diffusion Models (LDMs) provide both these capabilities, enabling high-quality super-resolution and naturalistic diversity enhancements. We demonstrate that enhancing the realism and diversity of the distilled dataset using LDMs results in a significant performance boost across various dataset distillation benchmarks.  

As LDMs become faster and more accessible \cite{podell2023sdxl}, the latency and computational costs of using them in dataset distillation are decreasing, enabling on-the-fly image augmentation. This trend lowers the barrier to incorporating LDMs, making their use more practical. Moreover, as LDMs become increasingly common in machine learning workflows, it's reasonable to expect that the student model could also leverage a diffusion model, ensuring consistent processing in teacher-student setups. This alignment makes LDM-enhanced distillation methods more feasible and appealing as LDMs continue to improve in efficiency and accessibility.

Our experiments across multiple datasets and model architectures demonstrate that distilling a dataset into a small set of images—such as one image per category—and training a student model on this distilled set results in higher accuracy than state-of-the-art dataset distillation methods. For example, on the ImageNette \cite{imagenette2019} dataset using a ResNet-18 architecture, our approach achieves 51.4\% accuracy, while RDED \cite{sun2024diversity}, a recent comparable baseline, reaches only 35.8\% accuracy.


Our specific contributions in this paper are as follows: 
\begin{enumerate}
    \item Using fast latent diffusion models (LDM) for on the fly coreset expansion in dataset distillation.        
    \item Employing knowledge distillation with generative models for dataset distillation.
    \item Significantly advancing state-of-the-art performance in large-scale dataset distillation.
\end{enumerate}

\begin{figure*}[t!]
    \centering
    \includegraphics[width=\linewidth]{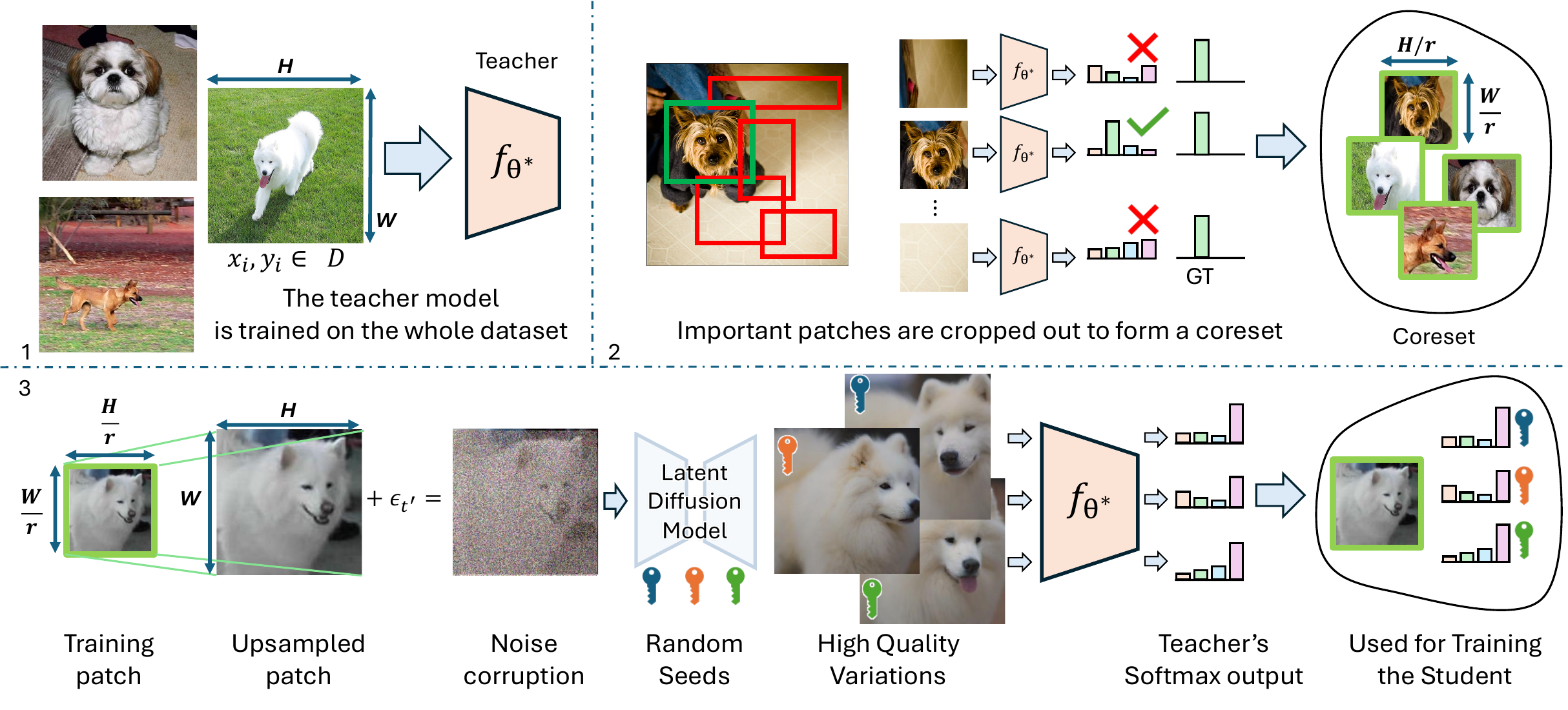}
    \caption{Proposed framework illustration: starting with an image dataset \( \mathcal{D} \), a teacher model is trained on the image-label pairs. Leveraging the uncertainty signal from the teacher's logits, following \cite{sun2024diversity}, we identify the most important patch from each image to form a coreset. These patches are then upsampled, noise-corrupted using fixed random seeds, and processed through a multi-step diffusion model to achieve simultaneous super-resolution and introduce variations to the coreset. For each random seed and generated high-resolution image, the teacher's soft label is obtained. The student then uses these important patches and random seeds to recreate the high-resolution images and regress over the teacher's corresponding soft labels. Note that, similar to traditional geometric augmentation techniques, this super-resolution and augmentation process is performed on the fly and discarded once the student's gradient is computed.}
    \label{fig:main}
\end{figure*}

\section{Related Works}

Since its introduction by \citet{wang2018dataset}, numerous variations of dataset distillation have been developed. Below, we review some of these methods as well as other related works to our proposed framework.

\noindent\textbf{Bi-level optimization} provides a natural framework for formalizing dataset distillation \cite{wang2018dataset}. However, as previously mentioned, this approach is generally intractable due to the significant computational and memory demands required to backpropagate through the unrolled computational graph of the inner model optimization. Various methods have been proposed to ameliorate this issue by introducing surrogate objective functions for the outer optimization. These include methods that match gradients \cite{zhao2021dataset,kim2022dataset}, extracted features \cite{wang2022cafe} and their distributions \cite{Zhao_2023_WACV}, and training trajectories \cite{cazenavette2022dataset,du2023minimizing}.

\noindent\textbf{Core-set selection} offers a natural approach to dataset distillation by selecting the most ``valuable" subset of the training data rather than synthesizing a condensed version. Core-set-based methods vary primarily in the difficulty-based metrics they use to evaluate sample importance. For example, sample scores such as Gradient Normed (GraNd) and Error L2 Norm (EL2N) were introduced in \cite{paul2021deep} to guide core-set selection. Meanwhile, the forgetting score, initially proposed in \cite{tonevaempirical}, has recently been employed to perform dataset distillation progressively \cite{chen2024data}. Recent works extend the concept of core-set selection by selecting a subset of pixels, tokens, or patches within chosen images, achieving further dataset compression \cite{zhou2023dataset, sun2024diversity}. For instance, \citet{sun2024diversity} select important image patches identified by a teacher model, while \cite{zhou2023dataset} use a subset of image tokens or patches and apply Masked Auto-Encoders \cite{he2022masked} to reconstruct the missing patches, resulting in dataset distillation conditioned on a generative model. In our work, we use important patches similar to \cite{sun2024diversity} but leverage generative modeling, and more precisely LDMs, to increase diversity and maximize compression, similar to \cite{zhou2023dataset}.

\noindent\textbf{Realism priors} have been explored in several studies to enhance the realism of distilled data. For example, \citet{cazenavette2022dataset} show that overfitting to a specific architecture often stems from optimizing in the pixel space, which reduces realism. They address this by using generative models—specifically, Generative Adversarial Networks (GANs), to perform optimization in the latent space, producing a more realistic distilled dataset leading to better cross-architecture generalization. Other approaches use a trained teacher model and align the feature statistics of synthetic distilled data with those of a larger dataset \cite{yin2024squeeze,shao2024generalized}, which implicitly improves image realism. In our work, we leverage generative priors to enhance both the realism and diversity of the distilled dataset.


\noindent\textbf{Diffusion models} have recently gained prominence as powerful tools for data augmentation and generation across various learning tasks \cite{sariyildiz2023fake, zhang2023expanding, shipard2023diversity, islam2024diffusemix, trabucco2024effective}. For example, \citet{zhang2023expanding} combine diffusion models with MAEs to expand small-scale datasets by generating new, informative, and diverse images, effectively creating realism-aware augmentations of limited datasets. DiffuseMix \cite{islam2024diffusemix} utilizes diffusion models and introduces a unique approach that blends real and generated images, producing hybrid augmentations. However, due to the slow generation speed of diffusion models, these augmentations are often pre-generated and cached, leading to high memory demands. With growing interest in diffusion models and advances in fast sampling techniques, such as SDXL-Turbo \cite{sauer2025adversarial}, it is now feasible to generate on-the-fly augmentations during model training. In our work, we leverage SDXL-Turbo to super-resolve and augment our mined, important patches.

\section{Method}

Our proposed method for dataset distillation comprises three main steps. In Step 1, a teacher model is trained on the full training dataset. In Step 2, a compact coreset of important image patches is selected. In Step 3, the selected image patches are first upsampled and then noise-corrupted using different fixed random seeds. Each seed generates a distinct high-quality variation of the low-resolution patch using a Latent Diffusion Model (LDM). Multiple such variations are generated for each patch using different seeds. The high-quality images are then processed by the teacher model to obtain the softmax outputs of its classifier. Finally, the low-quality important patches, random seeds, and their corresponding soft labels are transferred to the student model. Figure \ref{fig:main} demonstrates these steps. 

Upon receiving the distilled data from the teacher, the student replicates the upsampling, noise corruption, and LDM denoising steps using the provided low-quality patches and random seeds to generate the same high-quality variations as produced by the teacher. The student then trains on the teacher's soft labels for the generated images.


\subsection{Coreset Selection}



\begin{figure}[t!]
    \centering
    \includegraphics[width=\columnwidth]{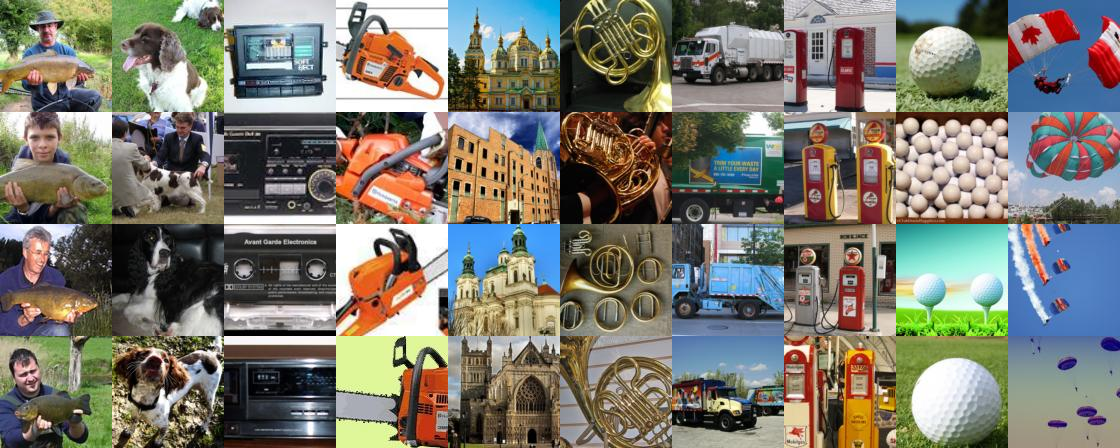}
    \caption{The extracted coreset for IPC=1 from ImageNette.}
    \label{fig:coreset}
\end{figure}

We follow the methodology of \cite{sun2024diversity} for forming the coreset of important patches. Given $\mathcal{D} = \{(x_i, y_i)\}_{i=1}^N$, where $x_i \in \mathbb{R}^{H \times W \times 3}$ are the images and $y_i \in \mathbb{R}^K$ are the corresponding class labels, we first train a teacher model $f_\theta:\mathbb{R}^{H \times W \times 3} \to \mathbb{R}^K$ parameterized by $\theta$ on $\mathcal{D}$. Then, for a given image $x_i$ from the dataset, we generate $P$ random crops and resize them to be $\lfloor\frac{H}{r}\rfloor\times\lfloor\frac{W}{r}\rfloor$, where $r>1$ is a scalar indicating the patch-to-image compression ratio. Denoting $x_i^j$ as the $j$-th random patch from $x_i$, we deliberately use a smaller patch size compared to the full dimensions of $x_i$ to compress the information, i.e., $x_i^j \in \mathbb{R}^{\lfloor H/r \rfloor \times \lfloor W/r \rfloor \times 3}$. To construct our coreset, we first select the most informative patch from each image. This is achieved by choosing the patch \( x^j_i \) that minimizes the cross-entropy loss:
\begin{equation}
    x^*_i = \underset{j}{\arg\min} \ \text{CE}(f_\theta(x_i^j), y_i),
\end{equation}
where $j=1,2,\dots,P$, and $\text{CE}(\cdot,\cdot)$ denotes the cross entropy loss. Next, to create the coreset under a fixed Images Per Class (IPC) memory budget, we form the coreset by choosing $P$ patches with the lowest cross entropy loss. At the end of this step, for each class \( c \), we will have a total of \( IPC \times r^2 \) patches, satisfying the memory constraint. Figure \ref{fig:coreset} shows the selected coreset for the ImageNette dataset \cite{imagenette2019} for IPC=1. We note that increasing \( r \) allows us to store a greater number of important patches for a fixed IPC budget, thereby enhancing diversity. However, this comes at the cost of reduced realism due to the loss in resolution.  Finally, we acknowledge that selecting patches based on minimum cross-entropy does not inherently ensure diversity. However, our results demonstrate that the diffusion model effectively compensates for any potential lack of diversity among the selected important patches. In the next step, we will describe our method for increasing realism despite increasing $r$.

\subsection{Coreset Augmentation and Super Resolution}
Data augmentation has long been a cornerstone for introducing variations into training data. In vision applications, simple and efficient geometric transformations such as rotations, flips, and noise additions are commonly used to artificially increase the dataset size. These augmentations are computationally inexpensive and can be performed on the fly. With recent advancements in Latent Diffusion Models (LDMs) \cite{sauer2025adversarial}, the development of few-step LDMs has significantly reduced the sampling latency traditionally associated with diffusion models. In this work, we leverage this advancement to perform realistic augmentations on important patches on the fly. Realistic variations are generated dynamically during the student model's batch processing and discarded once the batch is processed, ensuring compliance with memory storage constraints on the client side.

\subsubsection{Diffusion Preliminary}

Latent Diffusion Models (LDMs) consist of an autoencoder and a UNet denoiser. The autoencoder is an encoder-decoder architecture that is initially trained separately to map the image from pixel space to a lower-dimensional latent space and then reconstruct it back to the original image space with minimal reconstruction error. Once the autoencoder is trained, all noise corruptions and denoising are performed in the latent space. Let $z_0=\text{Encode}(x)$ be the latent representation of image $x$, and let $t \in \{0, 1, 2, \dots, T\}$ represent an arbitrary noising step. For training an LDM, $t$ is uniformly sampled from the set of steps, and Gaussian noise is added to $z_0$ in proportion to $t$. The noisy latent representation is given by:
\[
    z_t = \sqrt{\Bar{\alpha}_t}z_0 + \sqrt{1-\Bar{\alpha}_t}\epsilon,
\]
where $\sqrt{\Bar{\alpha}_t}$ is the data-noise interpolation coefficient and $\epsilon \sim \mathcal{N}(0, I)$. The denoiser, denoted by $\epsilon_\theta$, is then trained by minimizing the following loss function:

\[
   \mathcal{L}_{ldm} = \|\epsilon_\theta(z_t, t, c) - \epsilon\|^2
   \label{eq:ldm_loss}
\]
Here, $c$ is the conditioning vector. In the case of a text-to-image diffusion model, $c$ is the output of a text encoder that predicts the textual embedding of the image caption.

\subsubsection{Using real Data as Anchors}

Several works in the literature have investigated the potential of synthetic images for training downstream classifiers \cite{trabucco2023effective, he2022synthetic, fan2024scaling}. A consistent finding across these studies is that synthetic images can improve classifier performance when augmented with real data. However, even state-of-the-art synthetic images exhibit a slight distribution shift when not anchored to real images \cite{sariyildiz2023fake,fan2024scaling}. Consequently, real data and synthetic samples augmented from real data tend to outperform purely synthetic samples.

\begin{figure*}[t]
    \centering
    \includegraphics[width= \linewidth]{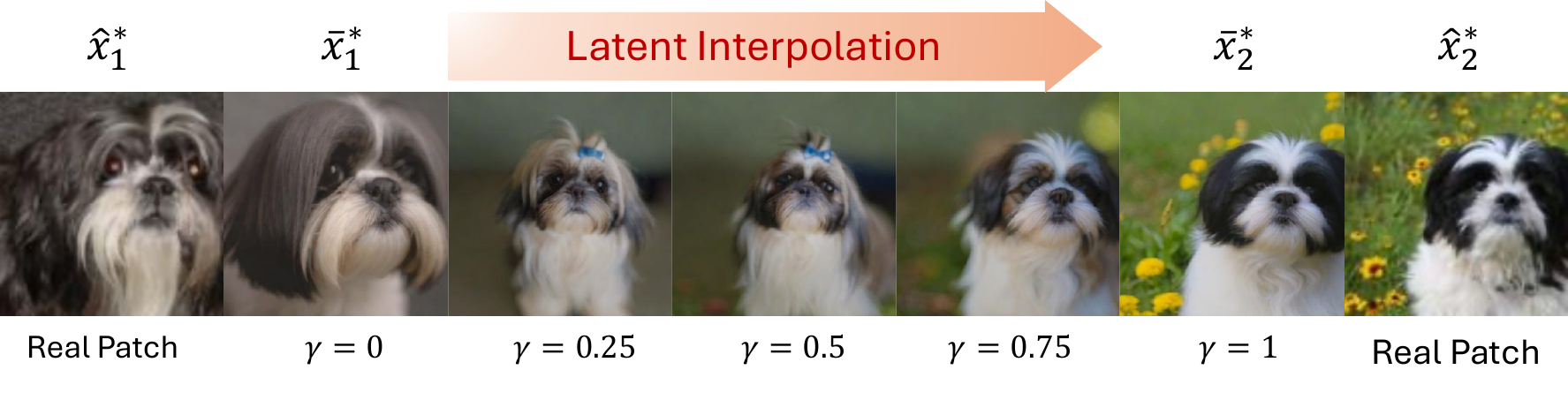}
    \caption{Performing mixup in the latent space of the LDM's autoencoder. }
    \label{fig:mixup_plot}
\end{figure*}

In this work, we propose to use the low-resolution patches stored during the coreset selection step as anchors on the manifold of training distribution. Let $x_i^* \in \mathbb{R}^{\lfloor H/r \rfloor \times \lfloor W/r \rfloor \times 3}$ be the $i^{\text{th}}$ low-resolution patch subsampled in the previous step. We propose to first upsample the patch using 2D interpolation to the original image dimensions:
\[
    \hat{x}_i^* = \text{INTERP}(x_i^*),
\]
where $\hat{x}_i^* \in \mathbb{R}^{H \times W \times 3}$. The upsampled patch matches the original image dimensions in $\mathcal{D}$ but still retains low quality. We propose using this low-quality image as an anchor for the LDM. Let $\hat{z}_i^*$ be the latent code of the upsampled patch. We first add a small amount of noise to $\hat{z}_i^*$:
\begin{equation}    
    \hat{z}_{i, t'}^* = \sqrt{\Bar{\alpha}_{t'}} \hat{z}_i^* + \sqrt{1-\Bar{\alpha}_{t'}}\epsilon,
\end{equation}
where $t'\in\{1, 2, \dots, T\}$. We define $\rho$ as $\frac{t'}{T} \in [0, 1]$ and use it as a hyperparameter that controls the amount of noise. This small noise addition perturbs the data slightly off the training data distribution manifold. We then iteratively denoise the low-quality latent back to the manifold using $\epsilon_\theta$, following the backward diffusion process:
\[
    \hat{z}_{i, {t'-1}}^* = \frac{1}{\sqrt{\alpha_{t'}}} \left( \hat{z}_{i, t'}^* - \frac{1-\alpha_{t'}}{\sqrt{1-\Bar{\alpha}_{t'}}} \epsilon_\theta(\hat{z}^*_{i, t'}, t', c) \right) + \sigma_t \epsilon'
\]
where $\epsilon' \sim \mathcal{N}(0, I)$, and $\Bar{\alpha}_{t'} = \prod_{s=1}^{t'} \alpha_{s}$. The denoised sample is denoted as $\Bar{z}_{i}^* = \hat{z}_{i, 0}^*$.

The denoised image $\Bar{x}_i^*=\text{Decode}(\Bar{z}_i^*)$ possesses two key properties: 1) since the denoiser is trained on high-resolution images, the denoised image will also be high-resolution, making $\Bar{x}_i^*$ high quality; and 2) as a projection onto the manifold of the training distribution, $\Bar{x}_i^*$ does not necessarily recover the same anchor patch, with the contents of $\Bar{x}_{i}^*$ varying slightly based on $c$ and $\rho$. Therefore, the final transformation results in a combination of super-resolution and semantic augmentation.

\subsection{Mixup in Latent Space}

Mixup \cite{zhang2018mixup} has become a widely adopted data augmentation method for training vision models. It encourages local linearity in the model by enforcing that the linear mixture of input images corresponds to the linear mixture of their outputs. This concept was later extended to manifold mixup \cite{verma2019manifold}, which better aligns with the underlying data manifold. In our approach, we assume that both the student and teacher models have access to an expressive LDM. This allows us to leverage the LDM to perform mixup operations in the latent space, effectively implementing manifold mixup, to further augment the student's limited set of samples. Let $\hat{z}_1^*$ represent the latent code of an upsampled patch in the training data belonging to class $k$. To augment this patch, we randomly sample another data point $\hat{z}_2^*$ from the same class and perform linear interpolation in the latent space with a mixing parameter $\gamma$, defined as $\hat{z}_{\text{interp}} = \gamma \hat{z}_1^* + (1-\gamma) \hat{z}_2^*$. The interpolated latent code is then used for augmentation via the LDM. Figure \ref{fig:mixup_plot} presents qualitative results of the augmented samples generated using mixup. In our ablation study, we highlight the performance improvements achieved by mixing latent codes.


\subsection{Putting it all together}

For each IPC, we extract $r^2$ low-resolution patches from our coreset selection. For each patch, we generate $m$ high-quality augmentations using the diffusion model and pass them through the teacher model to obtain soft labels, resulting in a total of $m \times r^2$ soft labels. In all our experiments, we set $m$ equal to the number of training epochs for the student. Notably, regenerating the high-quality augmentations requires only a single random seed. Lastly, we emphasize that the storage overhead for soft labels is present in many of the recent methods that combine knowledge distillation and dataset distillation, such as RDED \cite{sun2024diversity}, SRe2L \cite{yin2024squeeze}, and G-VBSM \cite{shao2024generalized}.

\begin{table*}[t]
\centering
\begin{tabular}{ccccccc}
\hline
          & & MTT & SRe2L & G-VBSM & RDED & Ours \\ 
          & & ConvNet & ResNet-18 & ResNet-18 & ResNet-18 & ResNet-18 \\ \hline
\multirow{3}{*}{Tiny-ImageNet} & IPC=1 & 8.8 $\pm$ 0.3 & 2.62 $\pm$ 0.1 & --- &  9.7 $\pm$ 0.4 &  19.4 \\ 
                               & IPC=10 & 23.2 $\pm$ 0.2 & 16.1 $\pm$ 0.2 & --- & 41.9 $\pm$ 0.2 & 46.2 \\ 
                               & IPC=50 & 28.0 $\pm$ 0.3 & 41.1 $\pm$ 0.4 & 47.6 $\pm$ 0.3 & 58.2 $\pm$ 0.1 & 53.4 \\ \hline
\multirow{3}{*}{ImageWoof}     & IPC=1 & 28.6 $\pm$ 0.8 & 13.3 $\pm$ 0.5 & --- & 17.9 $\pm$ 1.0 & 25.0 \\ 
                               & IPC=10 & 35.8 $\pm$ 1.8 & 20.2 $\pm$ 0.2 & --- & 44.4 $\pm$ 1.8 & 47.6 \\ 
                               & IPC=50 & --- & 23.3 $\pm$ 0.3 & --- & 71.7 $\pm$ 0.3 & 77.1\\ \hline
\multirow{3}{*}{ImageNette}    & IPC=1 & 47.7 $\pm$ 0.9 & 19.1 $\pm$ 1.1 & --- & 35.8 $\pm$ 1.0 & 51.4 \\ 
                               & IPC=10 & 63.0 $\pm$ 1.3 & 29.4 $\pm$ 3.0 & --- & 61.4 $\pm$ 0.4 & 73.6 \\ 
                               & IPC=50 & --- & 40.9 $\pm$ 0.3 & --- & 80.4 $\pm$ 0.4 & 87.6 \\ \hline
\multirow{3}{*}{ImageNet-1k}   & IPC=1 & --- & 0.1 $\pm$ 0.1 & --- & 6.6 $\pm$ 0.2 & 13.9 \\ 
                               & IPC=10 & --- & 21.3 $\pm$ 0.6 & 31.4 $\pm$ 0.5 & 42.0 $\pm$ 0.1 & 52.1 \\ 
                               & IPC=50 & --- & 46.8 $\pm$ 0.2 & 51.8 $\pm$ 0.4 & 56.5 $\pm$ 0.1 &  54.9\\ \hline
\end{tabular}
 \caption{Results on higher-resolution benchmarks. We compared our method against three knowledge-distillation-based approaches and MTT as a bilevel optimization method. Blank cells for the MTT method indicate its lack of scalability, while for G-VBSM represent results not reported by the authors. The results demonstrate that our approach is either superior or on par with the baselines. For ImageNet-1k, ImageWoof, and ImageNette, we used $112 \times 112$ patches, while Tiny-ImageNet experiments utilized $32 \times 32$ patches. In all experiments, the teacher and student models share the same architecture. Cross-architectural analysis results are detailed in the ablation studies.}
\label{tab:comparison}
\end{table*}

\section{Experiments}

We evaluated our method against both knowledge-distillation-based and bilevel-optimization-based approaches across several high-resolution benchmarks:

\begin{enumerate} \item \textbf{Tiny-ImageNet \cite{le2015tiny}:} This dataset includes 200 classes of $64 \times 64$ images derived from the original ImageNet-1k dataset. For our experiments, we selected patches at a resolution of $32 \times 32$.

\item \textbf{ImageWoof and ImageNette \cite{imagenette2019}:} These datasets are 10-class subsets of ImageNet-1k, with an original resolution of $224 \times 224$. ImageWoof focuses on different dog breeds, while Imagenette covers a broad array of categories spanning animals and objects. To ensure comparability with RDED, we used a patch size of $112 \times 112$.

\item \textbf{ImageNet-1k \cite{deng2009imagenet}:} A comprehensive dataset containing 1000 classes of $224 \times 224$ images representing a diverse set of categories. Consistent with \cite{sun2024diversity}, we employed patches of $112 \times 112$.
\end{enumerate}

\noindent While our method demonstrates its core strength on high-resolution benchmarks that benefit from super-resolution capabilities, we also benchmarked on CIFAR-10 and CIFAR-100 \cite{krizhevsky2009cifar} to address the challenges bilevel methods face with high-resolution images and widely used ResNet architectures. For these datasets, we used $16 \times 16$ patches.

The following baseline methods, including both bilevel-optimization-based and knowledge-distillation-based approaches, were used for evaluation:

\begin{enumerate} \item \textbf{RDED \cite{sun2024diversity}:} The first method that synthesizes collages of important patches, selected based on the teacher model's cross-entropy loss.

\item \textbf{SRe2L \cite{yin2024squeeze}:} Leverages batch norm statistics of the teacher model to perform model inversion, facilitating the synthesis of diverse samples.

\item \textbf{G-VBSM \cite{shao2024generalized}:} Extending \cite{yin2024squeeze}, this approach leverages rich statistical information from batch norm layers of multiple pretrained teachers to synthesize data.

\item \textbf{MTT \cite{cazenavette2022dataset}:} The first approach to define the objective of outer-level optimization by matching the training trajectory of the student model to the expert's.

\item \textbf{IDM \cite{zhao2023improved}:} Proposes efficient data distillation through distribution matching between synthetic and real data.

\item \textbf{Tesla \cite{cui2023scaling}:} Simplifies gradient calculations for trajectory-matching-based methods, enhancing the computational efficiency of \cite{cazenavette2022dataset}.

\item \textbf{DATM \cite{guo2023towards}:} This work improves the trajectory matching methods and aligns the complexity of generated patterns to the dataset's size.
\end{enumerate}

\vspace{0.1pt}

\vspace{0.1pt}

\noindent \textbf{Implementation Details:} We conducted all experiments using the float16 variant of the SDXL-Turbo diffusion model, while setting num\_inference\_steps=5. For most experiments, we employed the AdamW optimizer with a learning rate of 0.001 and a weight decay of 0.01, training for 300 epochs. Detailed descriptions of each experimental setup and the corresponding hyperparameters are provided in the supplementary material.

\begin{table*}[h!]
\hspace{-.7cm}
\small
\setlength{\tabcolsep}{3pt} 
\begin{tabular}{@{}cccccc|ccccccc@{}}
\hline
      &     & Tesla & MTT & IDM & DATM & G-VBSM & RDED & Ours & SRe2L & G-VBSM & RDED & Ours \\ 
       &    & ConvNet & ConvNet & ConvNet & ConvNet & ConvNet & ConvNet & ConvNet & ResNet18 & ResNet18 & ResNet18 & ResNet18 \\ \hline
\multirow{3}{*}{CIFAR10} & IPC=1 & 48.5 & 46.3 & 45.6 & 46.9 & --- & 23.5 & 38.2 & 16.6 & --- & 22.9 &31.0 \\ 
                         & IPC=10 & 66.4 & 65.3 & 58.6 & 66.8 & 46.5 & 50.2 & 64.56 & 29.3 & 53.5 & 37.1 &47.7 \\ 
                         & IPC=50 & 72.6 & 71.6 & 67.5 & 76.1 & 54.3 & 68.4 & 73.9 & 45.0 & 59.2 & 62.1 &70.4 \\ \hline
\multirow{3}{*}{CIFAR100} & IPC=1 & 24.8 & 24.3 & 20.1 & 27.9 & 16.4 & 19.6 & 42.63 & 6.6 & 25.9 & 11.0 &31.2 \\ 
                          & IPC=10 & 41.7 & 40.1 & 45.1 & 47.2 & 38.7 & 48.1 & 49.0 & 31.6 & 59.5 &42.6 & 57.7 \\ 
                          & IPC=50 & 47.9 & 47.7 & 50.0 & 55.0 & 45.7 & 57.0 & 53.5 & 50.2 & 65.0 & 62.6 &62.2 \\ \hline
                          
\end{tabular}
\caption{Results on CIFAR-10 and CIFAR-100 datasets. We evaluated our method against various bilevel-optimization and knowledge-distillation-based approaches using ResNet-18 and ConvNet-3 as the student/teacher architectures. For these low-resolution benchmarks, we stored and communicated $16 \times 16$ patches prior to super-resolution and augmentation. The results show that our method performs comparably to or better than the knowledge-distillation-based approaches. In all experiments, the student and teacher architectures were identical.}
\label{tab:comparison}
\end{table*}

\subsection{Effect of Patch Size}

RDED \cite{sun2024diversity} proposes compressing visual information by utilizing patches smaller than the original image dimensions. To explore the performance dynamics associated with varying patch sizes under a fixed memory budget, we conducted a study where, limited to storing one $224 \times 224$ image per class, we evaluated performance as patch sizes decreased and their number increased. In the $IPC=1$ scenario, we stored $r^2$ patches of size $\frac{H}{r} \times \frac{W}{r}$, with $r$ ranging from 2 to 8. For this analysis, we selected ImageWoof and Imagenette as benchmarks due to their differing classification dynamics: ImageWoof consists of 10 dog breeds with high inter-class similarity, while Imagenette includes 10 widely diverse categories with minimal inter-class relationship.

\begin{figure}[t!]
    \centering
    \hspace*{-0.8cm}
    \includegraphics[width=1.13\columnwidth]{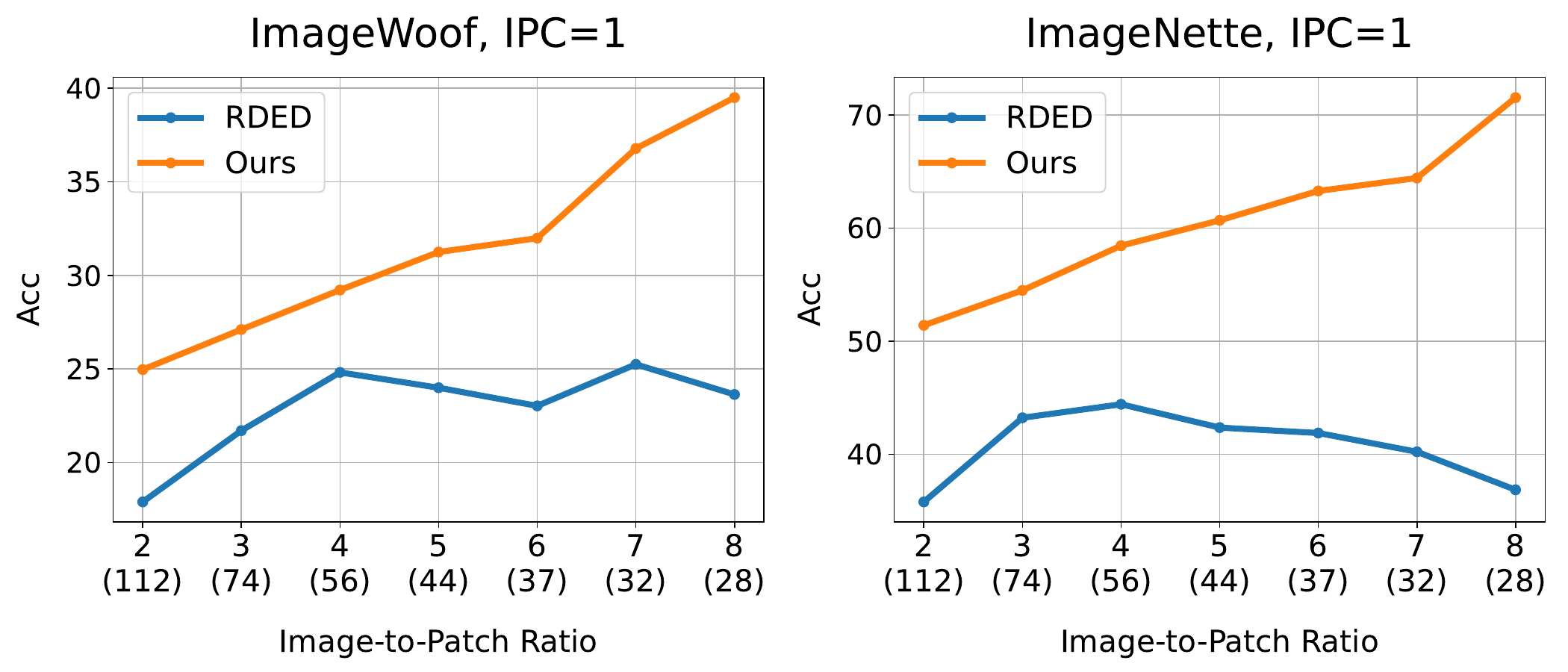}
    \caption{Study on the impact of patch size on student performance. In the RDED case, there is a trade-off between realism and diversity: reducing patch dimensions allows for more patches to fit in memory but significantly lowers the quality of downsampled patches without super-resolution. In contrast, our method benefits from increased performance by adding more patches, albeit at the cost of additional computation due to diffusion model calls. The x-axis represents varying $r$ values, with the numbers in parentheses indicating the corresponding patch sizes ($\frac{224}{r}$).}
    \label{fig:patchsz_ablation}
\end{figure}

Figure. \ref{fig:patchsz_ablation} illustrates that in RDED, increasing the number of patches enhances diversity but reduces realism, revealing an optimal performance point at $r=4$. In contrast, our method demonstrates a different trend: we consistently outperform RDED at all values of $r$, and our performance continues to improve as the patch count increases. This is achieved through our use of a diffusion model for super-resolution, which not only preserves realism but also enhances diversity by introducing variations to the data, complementing the diversity gains from the increased patch count. However, this improved performance comes at the cost of additional diffusion calls, leading to increased training time. Table \ref{tab:epoch_time} presents the average elapsed time per training epoch across various patch sizes. 

\begin{table}[h!]
\small
\hspace{-.8cm}
\centering
\begin{tabular}{c|cccccc}
 & \textbf{sz112} & \textbf{sz74} & \textbf{sz56} & \textbf{sz44} & \textbf{sz32} & \textbf{sz28} \\ \hline
\begin{tabular}[c]{@{}c@{}}\textbf{Epoch time}\\ \textbf{(sec)}\end{tabular} & 2.53 & 2.76 & 6.39 & 8.23 & 16.58 & 22.49 \\ 
\end{tabular}
\caption{Keeping IPC=1, one can increase $r$ to reduce the patch size and increase number of patches. We show the training time for each epoch in seconds on four RTX A6000 GPUs while varying the patch size.
}
\label{tab:epoch_time}
\end{table}

\subsection{Effect of Super-resolution and Augmentation}
\begin{figure}[t]
\centering
    \includegraphics[width=\columnwidth]{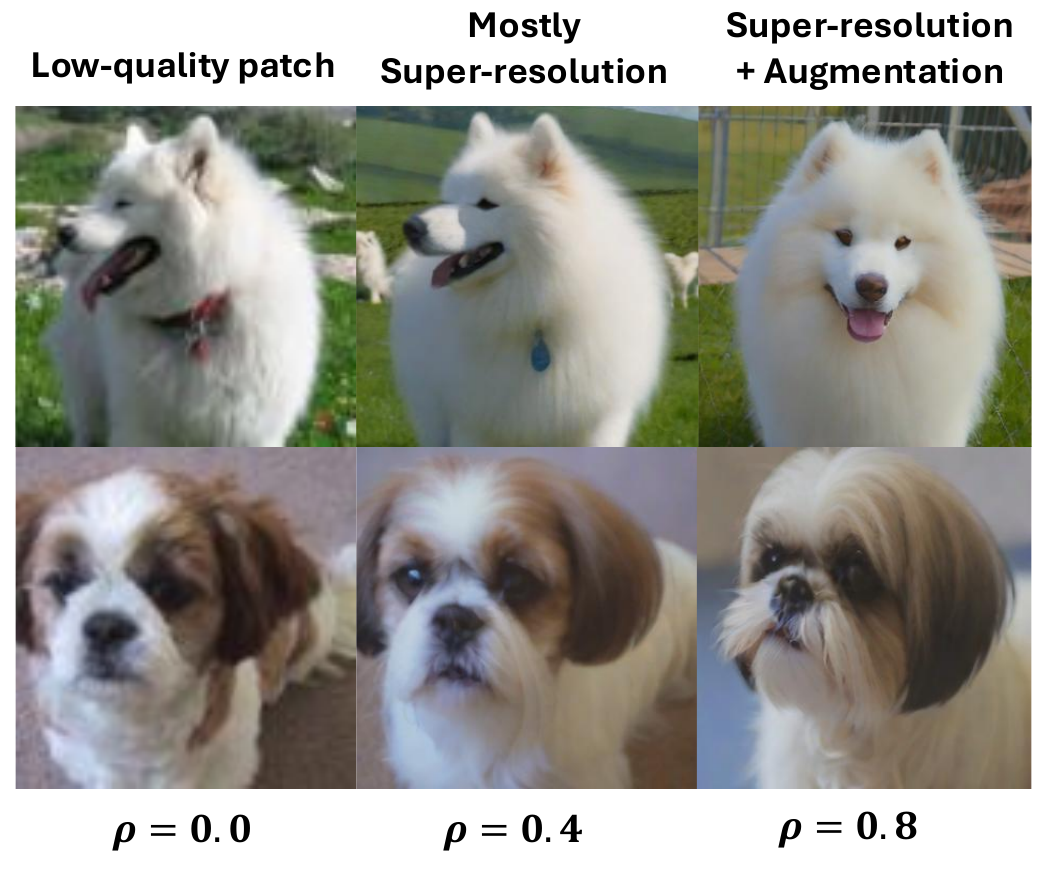}
    
    \caption{Qualitative illustration of the generated samples. By adjusting $\rho$, we can control the level of augmentation, allowing us to effectively distinguish between the contributions of super-resolution and augmentation.}

    \label{fig:rho_hyper}
\end{figure}

By denoising the latent code of a patch after adding partial noise, both content variation and super-resolution are achieved in the recovered image. In this study, we aim to disentangle these two operations to analyze their individual impact on accuracy. To simulate super-resolution with minimal augmentation, we set the ratio $\rho = \frac{t'}{T} = 0.4$, which we qualitatively observe to achieve super-resolution while preserving the original content. Conversely, at $\rho = 0.8$, additional variation is also introduced into the patches. Figure. \ref{fig:rho_hyper} presents qualitative results from this disentanglement.

In Table \ref{tab:aug_ablation}, we illustrate the contribution of various components to overall performance. In ``RDED," training is conducted exclusively on real patches. ``Only text cond." refers to training on synthetic data using only the textual prompt, ``A photo of \textit{category\_name}," without storing real patches. Setting $\rho=0.4$ generates high-quality samples with minimal augmentation, while $\rho=0.8$ enables both augmentation and super-resolution. The results indicate that the best performance is achieved when super-resolution, augmentation, and latent mixup are incorporated during training.


\begin{table}[h!]
\centering
\small 
\renewcommand{\arraystretch}{1.2} 
\setlength{\tabcolsep}{6pt} 
\begin{tabular}{c|c|c}
\hline
                                 & \textbf{ImageWoof} & \textbf{ImageNette} \\ \hline
RDED                             & 17.9               & 35.8                \\ \hline
Only text cond.                  & 19.6               & 35.1                \\ \hline
\begin{tabular}[c]{@{}c@{}}Superres\\ ($\rho = 0.5$)\end{tabular}   & 19.5               & 39.5                \\ \hline
\begin{tabular}[c]{@{}c@{}}Superres+Aug\\ ($\rho = 0.8$)\end{tabular} & 21.6               & 47.7               \\ \hline
\begin{tabular}[c]{@{}c@{}}Superres+Aug+\\ Mixup ($\rho = 0.8$)\end{tabular} & 25.0          & 51.4               \\ \hline
\end{tabular}
\caption{Ablation study to highlight the contributions of individual components in our framework and to separate the effects of super-resolution from augmentation. The experiments were performed on the ImageWoof and ImageNette datasets, with IPC=1 and using $112 \times 112$ patches. The student is a ResNet-18 model.}

\label{tab:aug_ablation}
\end{table}

\subsection{Cross-architectural Analysis}
Bilevel-optimization-based methods often struggle with poor cross-architectural transferability, meaning the performance of a student model significantly degrades when its architecture differs from that of the expert model used for dataset distillation. To address this, GLaD \cite{cazenavette2023generalizing} leverages the prior of a generative model to synthesize more realistic samples. \citet{cazenavette2023generalizing} demonstrate that this enhanced realism substantially improves the cross-architectural performance of distilled datasets. Since our method also utilizes the prior of a generative model, we compare its cross-architectural capabilities with those of GLaD in both high-resolution and low-resolution settings.

In Table \ref{tab:cross_arch_highres}, we report results using a ConvNet model as the expert/teacher in the IPC=1 setting. The values reflect the average performance across four different student architectures: VGG11 \cite{simonyan2014very}, ViT \cite{dosovitskiy2020image}, AlexNet \cite{krizhevsky2012imagenet}, and ResNet-18 \cite{he2016deep}. Since GLaD does not scale to the full resolution of $224 \times 224$, their ImageNette and ImageWoof images were downsampled to $128 \times 128$; we adjusted our setting accordingly to ensure a fair comparison. Table \ref{tab:tab:cross_arch_lowres} provides a similar cross-architectural analysis on CIFAR-10. Both tables show that our method achieves superior cross-architecture performance.

\begin{table}[h!]
\centering
\begin{tabular}{ccc}
\hline
           & Imagenette & ImageWoof \\ \hline
MTT \cite{cazenavette2022dataset}         & 24.1               & 16.0                \\ 
MTT + \textbf{GLaD}     & 30.4               & 17.1                \\ \hline
DC \cite{zhao2021dataset}          & 28.2               & 17.4                \\ 
DC + \textbf{GLaD}      & 31.0               & 17.8                \\ \hline
DM \cite{Zhao_2023_WACV}           & 20.6               & 14.5                \\ 
DM + \textbf{GLaD}      & 21.9               & 15.2                \\ \hline
\textbf{Ours}         & \textbf{30.79 }             & \textbf{19.74  }             \\ \hline
\end{tabular}
\caption{Cross-architectural analysis on ImageNette and ImageWoof at a resolution of $128 \times 128$ with IPC=1. The teacher model is a ConvNet, while the student architectures include VGG11, ViT, ResNet18, and AlexNet. The reported results represent the average performance across these architectures. In this setup, $64 \times 64$ patches were communicated to the student models within the memory constraints.}
\label{tab:cross_arch_highres}
\end{table}

\begin{table}[h!]
\centering
\begin{tabular}{ccccc}
\hline
           & AlexNet & ResNet18 & ViT & Average \\ \hline
MTT          & 26.8    & 23.4    & 21.2    & 23.8   \\ 
MTT + \textbf{GLaD}     & 27.9    & 30.2    & 22.7    & 26.9   \\ \hline
DC           & 25.9    & 27.3    & 22.9    & 25.4   \\ 
DC + \textbf{GLaD}      & 26.0    & 27.6    & \textbf{23.4}    & 25.7   \\ \hline
DM           & 22.9    & 22.2    & 21.3    & 22.1   \\ 
DM + \textbf{GLaD}      & 25.1    & 22.5    & 23.0    & 23.5   \\ \hline
\textbf{Ours}         & \textbf{31.3}    & \textbf{36.0}    & 21.9    & \textbf{29.7}   \\ \hline
\end{tabular}
\caption{Cross-architectural analysis on the CIFAR-10 dataset using a ConvNet as the teacher and various student architectures. The results demonstrate superior overall cross-architecture performance achieved by our method.}
\label{tab:tab:cross_arch_lowres}
\end{table}

\section{Conclusion}

Recent advancements in dataset distillation have underscored the significance of realistic and diverse data representations. Some approaches emphasize the value of realism for generalizability, while others explore the capabilities of generative models to enhance the diversity and quality of distilled datasets. Building on these insights, our proposed method leverages modern Latent Diffusion Models (LDMs) to address both realism and diversity. By combining coreset selection with generative augmentations, we achieve significant improvements in dataset distillation benchmarks, demonstrating state-of-the-art performance across various datasets. Our experiments validate the effectiveness of high-quality augmentations and mixup operations in the latent space, showcasing the power of LDMs to enhance dataset compression while preserving crucial data attributes.

{\small
\bibliographystyle{ieeenat_fullname}
\bibliography{11_references}
}


\end{document}

%% file: _constants.tex
\def\authorBlock{
    
    Ali Abbasi$^1$\thanks{~Work has been partly done during an internship at Microsoft Research} ~, Shima Imani$^2$\thanks{~Contributed to the project while working at Microsoft Research}, Chenyang An$^3$, Gayathri Mahalingam$^2$, \\ Harsh Shrivastava$^2$, Maurice Diesendruck$^2$\footnotemark[2], Hamed Pirsiavash$^4$, Pramod Sharma$^2$\thanks{~Equal contribution}~, Soheil Kolouri$^1$\footnotemark[3] \vspace{.4cm} \\  Vanderbilt University$^1$ \quad
Microsoft Research$^2$ \quad University of California, San Diego$^3$ \\ University of California, Davis$^4$ \vspace{.2cm} \\
  \texttt{\{ali.abbasi, soheil.kolouri\}@vanderbilt.edu}\\
  \texttt{c5an@ucsd.edu}\\
  \texttt{\{gmahalingam, hshrivastava, pramod.sharma\}@microsoft.com}\\
  \texttt{moglobal@gmail.com} ,~ \texttt{shimaimani@meta.com}
}

\newif\ifreview 
\newif\ifarxiv \newcommand{\arxiv}{\arxivtrue}
\newif\ifcamera 
\newif\ifrebuttal 

%% file: cvpr_header.tex
\ifreview \usepackage[review]{cvpr} \fi
\ifarxiv \usepackage[pagenumbers]{cvpr} \fi
\ifrebuttal \usepackage[rebuttal]{cvpr} \fi
\ifcamera \usepackage{cvpr} \fi

\input{_macros}  

\usepackage{xr-hyper}

\makeatletter
\newcommand*{\addFileDependency}[1]{
  \typeout{(#1)}
  \@addtofilelist{#1}
  \IfFileExists{#1}{}{\typeout{No file #1.}}
}

\makeatother

\definecolor{cvprblue}{rgb}{0.21,0.49,0.74}
\usepackage[pagebackref,breaklinks,colorlinks,allcolors=cvprblue]{hyperref}
\usepackage[capitalize]{cleveref}
\crefname{section}{Sec.}{Secs.}
\crefname{table}{Table}{Tables}
\crefname{figure}{Fig.}{Figs.}

\ifarxiv \crefname{appendix}{App.}{Apps.}
\else \crefname{appendix}{Suppl.}{Suppls.} \fi

%% file: _macros.tex
\usepackage{graphicx}	
\usepackage{amsmath}	
\usepackage{amssymb}	
\usepackage{booktabs}
\usepackage{times}
\usepackage{microtype}
\usepackage{epsfig}
\usepackage{caption}
\usepackage{float}
\usepackage{placeins}
\usepackage{color, colortbl}
\usepackage{stfloats}
\usepackage{enumitem}
\usepackage{tabularx}
\usepackage{xstring}
\usepackage{multirow}
\usepackage{xspace}
\usepackage{url}
\usepackage{subcaption}
\usepackage{xcolor}
\usepackage[hang,flushmargin]{footmisc}

\ifcamera \usepackage[accsupp]{axessibility} \fi

% ...

\ifarxiv  \fi

\newcommand{\R}[1]{{%
    \textbf{%
        \ifstrequal{#1}{1}{\textcolor{red}{R#1}}{%
        \ifstrequal{#1}{2}{\textcolor{blue}{R#1}}{%
        \ifstrequal{#1}{3}{\textcolor{magenta}{R#1}}{%
        \ifstrequal{#1}{4}{\textcolor{teal}{R#1}}{%
                           \textcolor{cyan}{R#1}%
        }}}}%
    }%
}}